\documentclass[10pt,twocolumn,letterpaper]{article}

\usepackage{iccv}
\usepackage{times}
\usepackage{epsfig}
\usepackage{graphicx}
\usepackage{amsmath}
\usepackage{amssymb}
\DeclareMathOperator{\EX}{\mathbb{E}}
\newcommand{\Lagr}{\mathcal{L}}

\usepackage[breaklinks=true,bookmarks=false]{hyperref}

\iccvfinalcopy 

\def\iccvPaperID{****} 

\ificcvfinal\fi

\begin{document}

\title{Unsupervised Projection Networks for Generative Adversarial Networks}

\author{Daiyaan Arfeen\thanks{equal contribution} \ and Jesse Zhang\footnotemark[1]\\
UC Berkeley\\
{\tt\small \{daiyaanarfeen,jessezhang\}@berkeley.edu}}

\maketitle
\ificcvfinal\thispagestyle{empty}\fi
\begin{figure*}[htp]
  \centering
  \includegraphics[width=\textwidth]{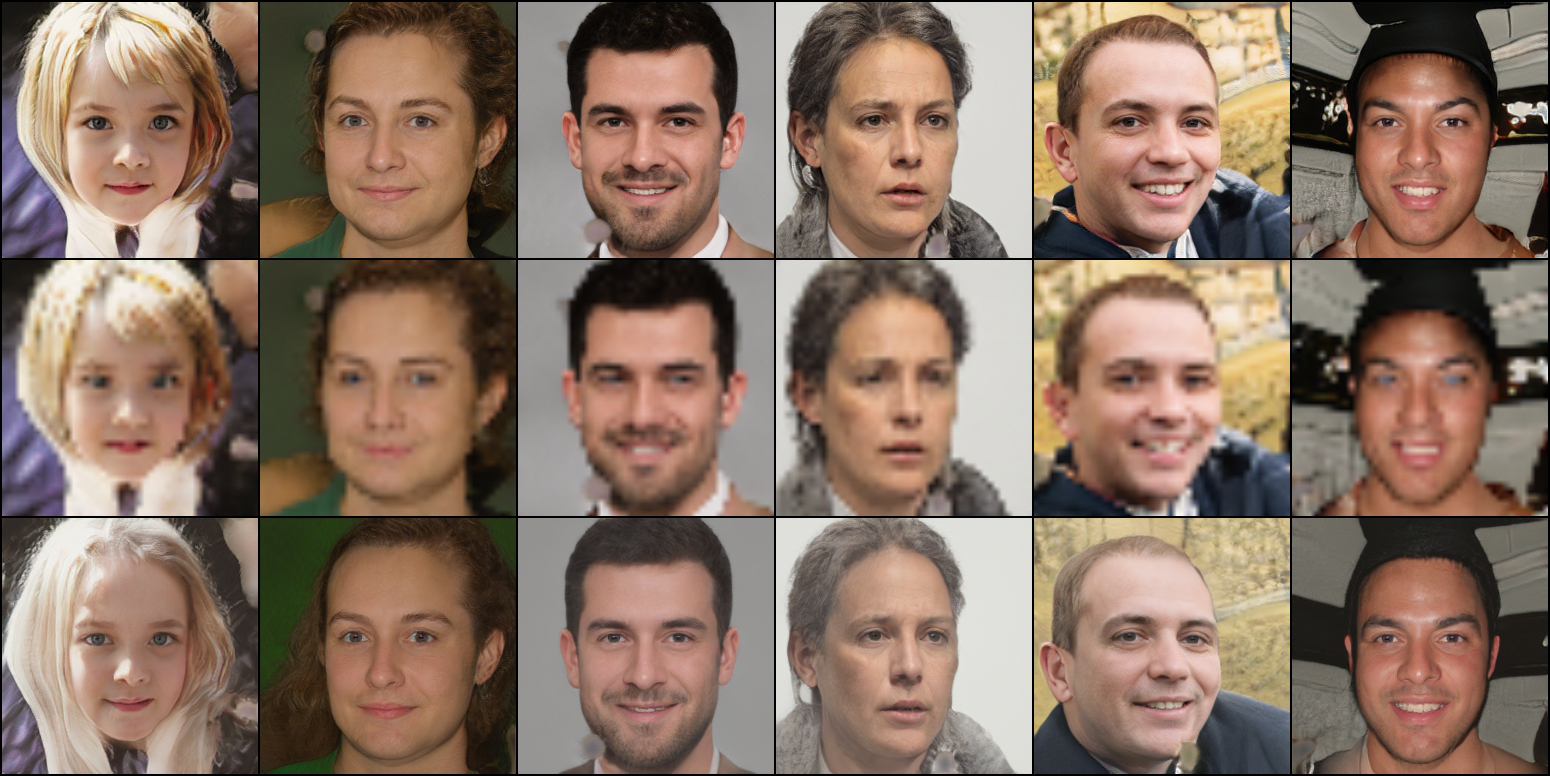}
  \caption{Example of image super-resolution performed by our model. Top: original images at $256 \times 256$. Middle: downsampled images at $45 \times 45$. Bottom: the resulting images upsampled back to $256 \times 256$}
  \label{front_page_figure}
\end{figure*}
\begin{abstract}
  We propose the use of unsupervised learning to train projection networks that project onto the latent space of an already trained generator. We apply our method to a trained StyleGAN, and use our projection network to perform image super-resolution and clustering of images into semantically identifiable groups. 
\end{abstract}

\section{Introduction}

Deep learning has been rapidly pushing the boundaries for image processing, generation and manipulation works. Generative Adversarial Networks (GANs) in particular have been achieved success in this domain \cite{oggan}. StyleGAN, a recently published network architecture for face generation \cite{stylegan}, is able to automatically separate high level features in an unsupervised manner while preserving stochasticity in the generated images. This generator can better disentangle complex image properties into latent factors, allowing for better control and interpolation properties. We propose in this paper a new method that constructs a network that maps an image to its latent code for the purpose of deriving structure.
    
In this paper, we explore whether the image to latent network can learn a projection onto a latent space that can be used to explore structure of the images; we demonstrate this by using it to cluster real images and perform image super-resolution. Our results show that the projection representation network is successful as we are able to achieve image super-resolution on images produced by the generator. Furthermore, utilizing the projection network for clustering results in meaningful, varied clusters of real images.

\section{Related Work}

Recently, deep generative models such as generative adversarial networks (GANs) have been popular in the image generation literature \cite{oggan}. GANs consist of generator and discriminator networks in which the generator’s goal is to produce synthetic images that trick the discriminator into believing that they are real. The training sequence resembles finding the optimum with two players having opposing objectives. When trained successfully, the two-player game ideally results in the generator being able to generate realistic images while the discriminator cannot do better than random guessing.


Since the original GAN paper, myriad GAN training techniques and network architectures have been developed to allow generators to better approximate high quality, varied image distributions. For example, StyleGAN \cite{stylegan} uses a learned fully-connected network to disentangle the randomly sampled latent space for separation of high-level image attributes. When trained on faces, this latent vector likely encodes both global and local information such as pose, gender, face shape, hair style, etc.

Even with a disentangled latent space, StyleGAN still suffers a common challenge faced by GANs: it is difficult to derive the structure of the latent space. This is problematic for the application of GANs to controlled image generation, as the latent space is the main source of variance for the final output. Clustering is one task that can expose structure in a space, allowing for partitioning the space into various diagrams. \cite{clustergan} showed good results clustering images using their projections into the latent space of a GAN, but their method requires training a network to learn a projection from the image space to the latent space as well as training the GAN with a new loss. \cite{ali} trained an encoder to learn the mapping from image to latent space jointly while training the generator and discriminator, while \cite{bigan} proposed a bidirectional GAN architecture that includes an inverting network and similarly projects images into the GANs latent space. Learned projections have also been proposed by \cite{efrosmanifold} where the projection functions was learned for an already trained GAN for the purpose of interpolation. 

In this paper we learn a projection function for an already trained StyleGAN so that we may suitably cluster images in the StyleGAN latent space. We do so by training the projection network with a latent loss in an unsupervised manner. We note that our technique should work for most already trained generator architectures; in this paper we explore StyleGAN exclusively.

\section{Methodology}
The StyleGAN generator consists of two networks that are jointly trained: a multi-layer perceptron (MLP) that projects an entangled latent variable $z\in Z$ drawn from a random distribution into a disentangled latent space $W$, and a generator that transforms a disentangled latent variable $w\in W$ into an image. We denote the MLP as $\mathbf{F}$ and the generator as $\mathbf{G}$. 

Our objective is to find an estimator $\mathbf{P}$, that projects images to a disentangled latent space, in order to minimize over $\Theta$, the parameter space, the risk taken over the latent variable $Z$. Our loss function is the squared error loss. Thus, the parameters of $\mathbf{P}$ are denoted as:
$$\theta^{*} = argmin_{\Theta} \EX_{z \sim Z} [\vert \vert \mathbf{P}(\mathbf{G}(\mathbf{F}(z))) - \mathbf{F}(z) \vert \vert_2^2 ]$$

If $\mathbf{G}$ is not StyleGAN's generator, $\mathbf{F}$ is usually the identity function and we would be projecting from images directly to the randomly sampled latent space $Z$.

Unlike previous approaches to learning a projection onto the latent space of a GAN, our method is unsupervised and only relies on the trained generator. Previous approaches, such as \cite{efrosmanifold}, used supervised learning to train the projection network such that the parameters of $\mathbf{P}$ are:
$$\theta^{*} = argmin_{\Theta} \EX_{x \sim X} [\Lagr(\mathbf{G}(\mathbf{P}(x)), x)]$$
where $X$ is the image dataset and $\Lagr$ is a loss function such as MSE or negative log-likelihood.  

\section{Experiments and Results}
\subsection{Training a Baseline Model}
\begin{figure*}[ht]
  \centering
  \includegraphics[width=\linewidth]{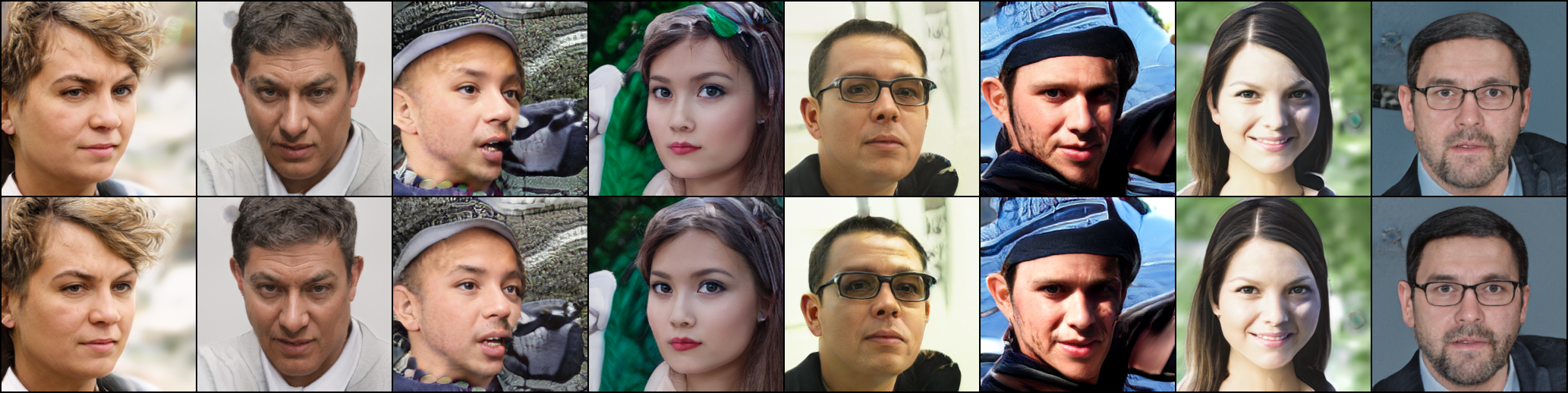}
  \caption{Top: Images randomly sampled from $\mathbf{G}$. Bottom: Images reconstructed by $\mathbf{G}$ using $\mathbf{P}$'s projection of the original images.}
  \label{10_minutes}
\end{figure*}

For all of our experiments, our network $\mathbf{P}$ is the ResNet18 model \cite{resnet} pretrained on ImageNet \cite{imagenet} with the fully connected layer replaced with a randomly initialized one. The generator $\mathbf{G}$ and the network $\mathbf{F}$ come from a pretrained StyleGAN model \cite{stylegan} (trained on the FFHQ Dataset) with output size $256 \times 256$, held constant during training of $\mathbf{P}$. We approximate $\theta^{*}$ through stochastic gradient descent optimized with the Adam optimizer. In less than 10 minutes of training on a single Tesla V100 GPU, $\mathbf{P}$ is able to learn a qualitatively good projection onto $\mathbf{G}$, as shown in Figure \ref{10_minutes}.
\subsection{Super-Resolution}
\begin{figure*}[ht]
  \centering
  \includegraphics[width=\linewidth]{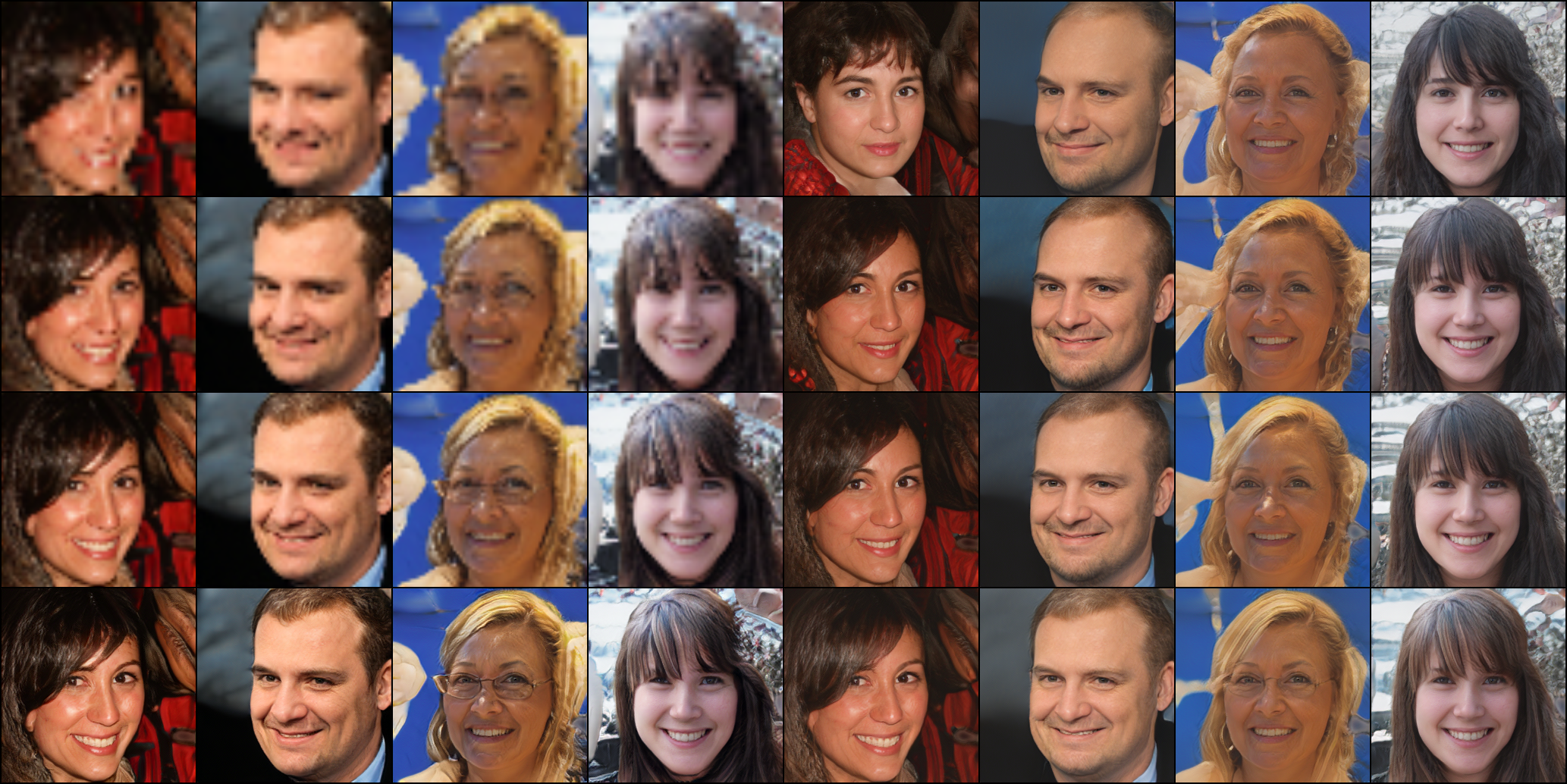}
  \caption{Comparing super-resolution results at different resolutions. In order from top to bottom, the left side displays images produced by $\mathbf{G}$ downsampled to $32 \times 32$, $45 \times 45$, $64 \times 64$, and $256 \times 256$ (no downsampling). The right side shows those same images as reconstructed by $\mathbf{P}$ and $\mathbf{G}$.} 
  \label{Resolution Comparison}
\end{figure*}

We observe that our projection network $\mathbf{P}$ compresses the most important information about the face in the input image. Even with an input image that is at a much lower resolution than what $\mathbf{P}$ is trained to project, the distinguishing details are still extracted from the face. Thus we utilize $\mathbf{P}$ trained on images at $256 \times 256$ to super-resolve images downsampled to a much lower resolution by feeding it the downsampled image and then feeding the projection into $\mathbf{G}$. This results in accurate recreations of images when downsampled to $64 \times 64$ and even $45 \times 45$, however the images are noticeably different from the others once downsampled to $32 \times 32$. A comprehensive comparison with the same examples downsampled to different resolutions is shown in Figure \ref{Resolution Comparison}.

\begin{figure}[ht]
  \centering
  \includegraphics[width=\linewidth]{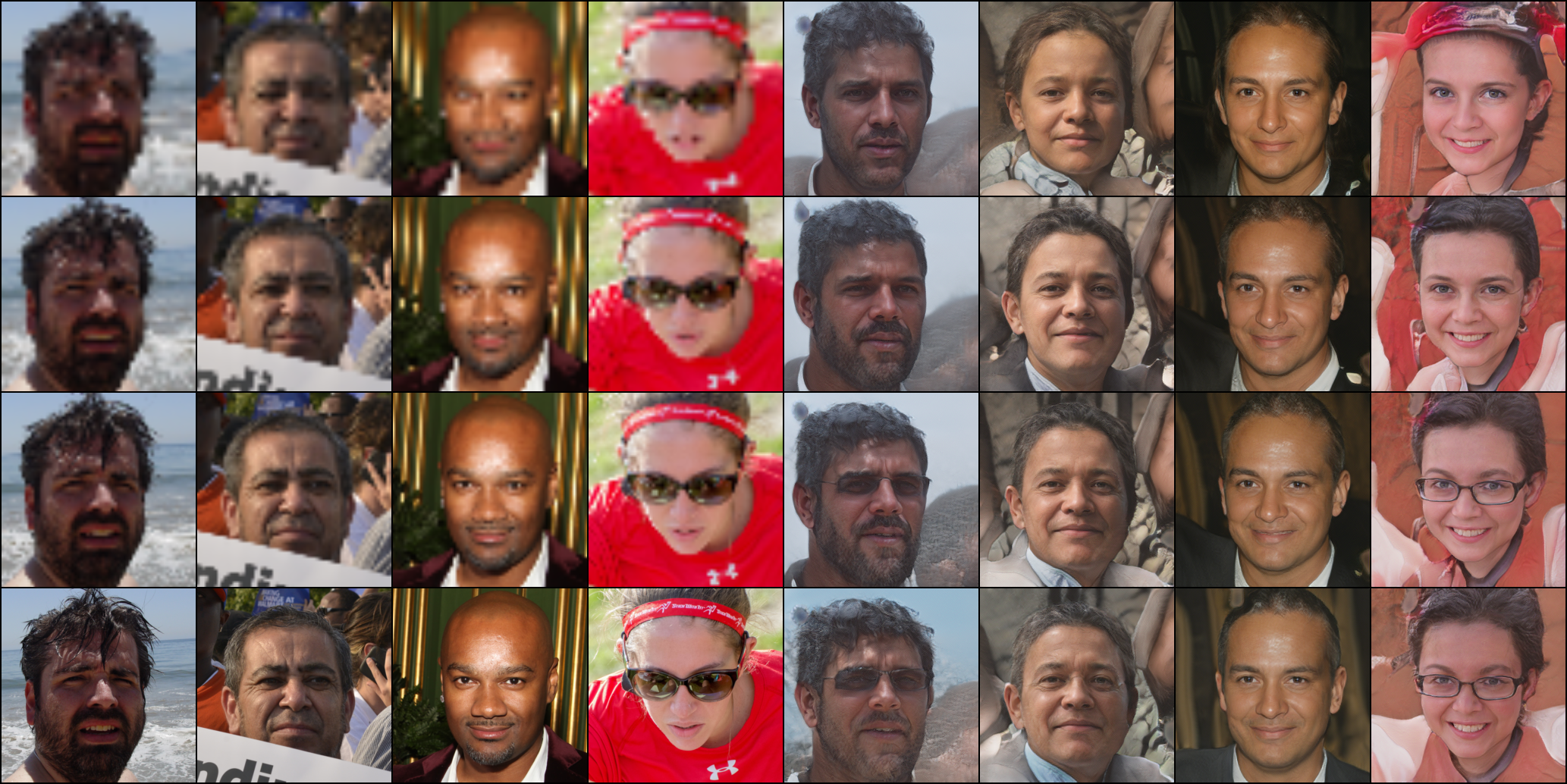}
  \caption{The same comparison as above but using images from FFHQ. Clearly $\mathbf{P}$ and $\mathbf{G}$ are unable to faithfully reproduce these images.}
  \label{FFHQ Resolution Comparison}
\end{figure}

We also attempted to perform the same experiment with images from the FFHQ dataset directly, however the images produced are not similar enough to original images. An example is shown in Figure \ref{FFHQ Resolution Comparison}.

\subsection{Distribution Extension}
One consequence of training on images only drawn from the generator's distribution is the inability to extend to images not in the generator's output space. This can be seen as a form of overfitting and limits the application of our method. We attempt to alleviate this issue through two methods, described below.
\subsubsection*{Fine-tuning}
\begin{figure}[ht]
  \centering
  \includegraphics[width=\linewidth]{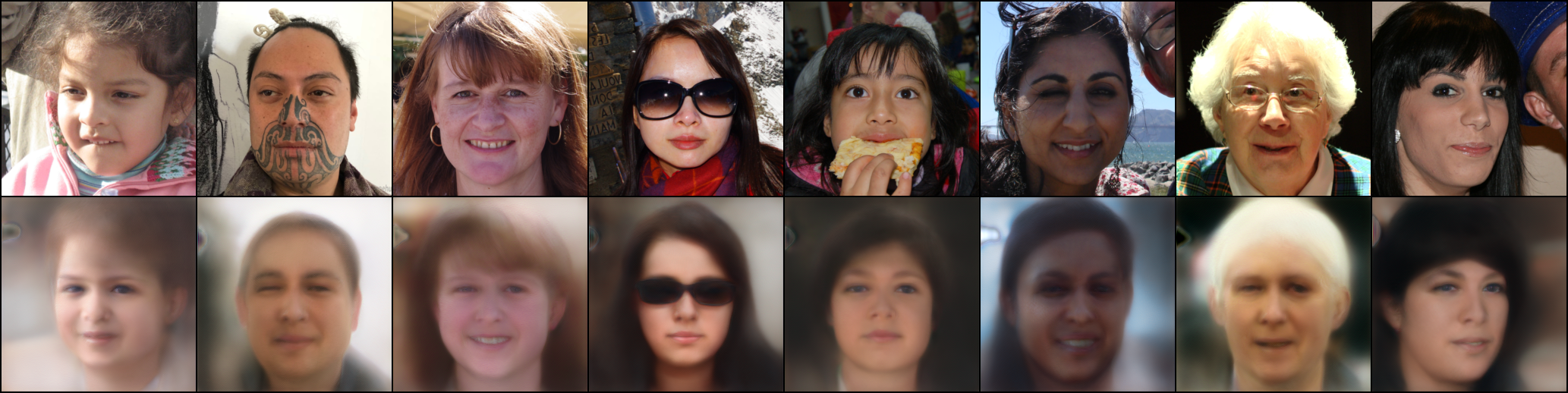}
  \caption{An example of the images that look too "smoothed" from the reconstruction loss.}
  \label{Fine Tuning}
\end{figure}

 After we train our network as described in section 4, we further train our network on a reconstruction loss. Specifically, we take an image $\mathit{I}_{FFHQ}$ from the FFHQ dataset, and we train to minimize the risk
 $$\EX_{x \sim X} [ \vert \vert x - \mathbf{G}(\mathbf{P}(x)) \vert \vert_2^2 ]$$ where $X$ is the set of FFHQ images.
 We continue training with the original loss, and we simply add our reconstruction loss, multiplied by a weighting factor, to the original loss. Unfortunately, we did not see improvements with this loss and instead saw our model regress to producing more "general" images that lack detail and look more "smoothed" as can be seen in Figure \ref{Fine Tuning}.
\subsubsection*{Jointly Training $\mathbf{G}$ and Utilizing a Discriminator}
\begin{figure}[ht]
  \centering
  \includegraphics[width=\linewidth]{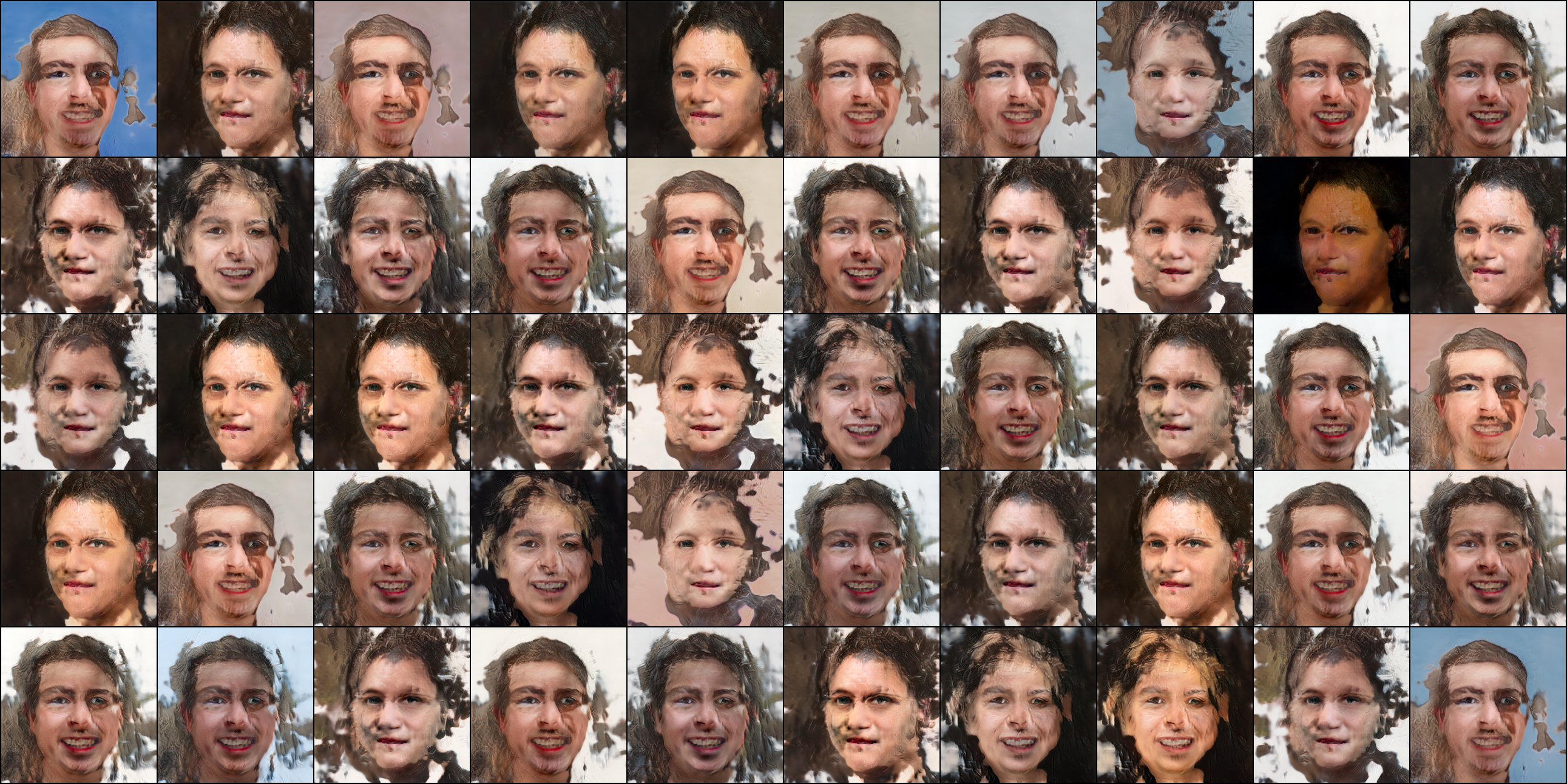}
  \caption{A example of the images produced when $\mathbf{P}$ and $\mathbf{G}$ fail to converge during training when we attempt to make them learn to reproduce images from FFHQ. We see definite mode collapse despite utilizing losses that are designed to prevent it.}
  \label{Convergence Failure}
\end{figure}

In order to avoid the smoothing described above, we utilize a discriminator to ensure that the images still look like faces in FFHQ dataset. We use a standard discriminator model with a series of downsampling convolutional layers connected to a final fully connected layer. We tried a variety of losses on the generator, including the original discriminator loss \cite{oggan}, a feature matching loss from pix2pixHD \cite{pix2pixhd}, and the standard discriminator loss utilized in pix2pixHD \cite{pix2pixhd}. All of these losses result in $\mathbf{P}$ being unable to converge during training, so we then allowed $\mathbf{G}$ to be trained jointly with $\mathbf{P}$. Even this did not converge during training, resulting in even worse pictures than before (see Figure \ref{Convergence Failure} for an example of what happens when $\mathbf{P}$ and $\mathbf{G}$ cannot converge). We also tried replacing the $MSE$ reconstruction loss with $VGG$ loss \cite{vgg}, similar to that of \cite{gatys2015}, however similar smoothing still occured. Ultimately, we believe that these methods are insufficient to allow $\mathbf{P}$ and $\mathbf{G}$ to reproduce images outside of the distribution that $\mathbf{G}$'s images are from, and furthe
r study is needed to determine whether it is possible at all to do so. For now, our $\mathbf{P}$ is limited to only properly encoding images from $\mathbf{G}$'s distribution.
\subsection{Image Clustering}
\begin{figure}[ht]
  \centering
  \includegraphics[width=\linewidth]{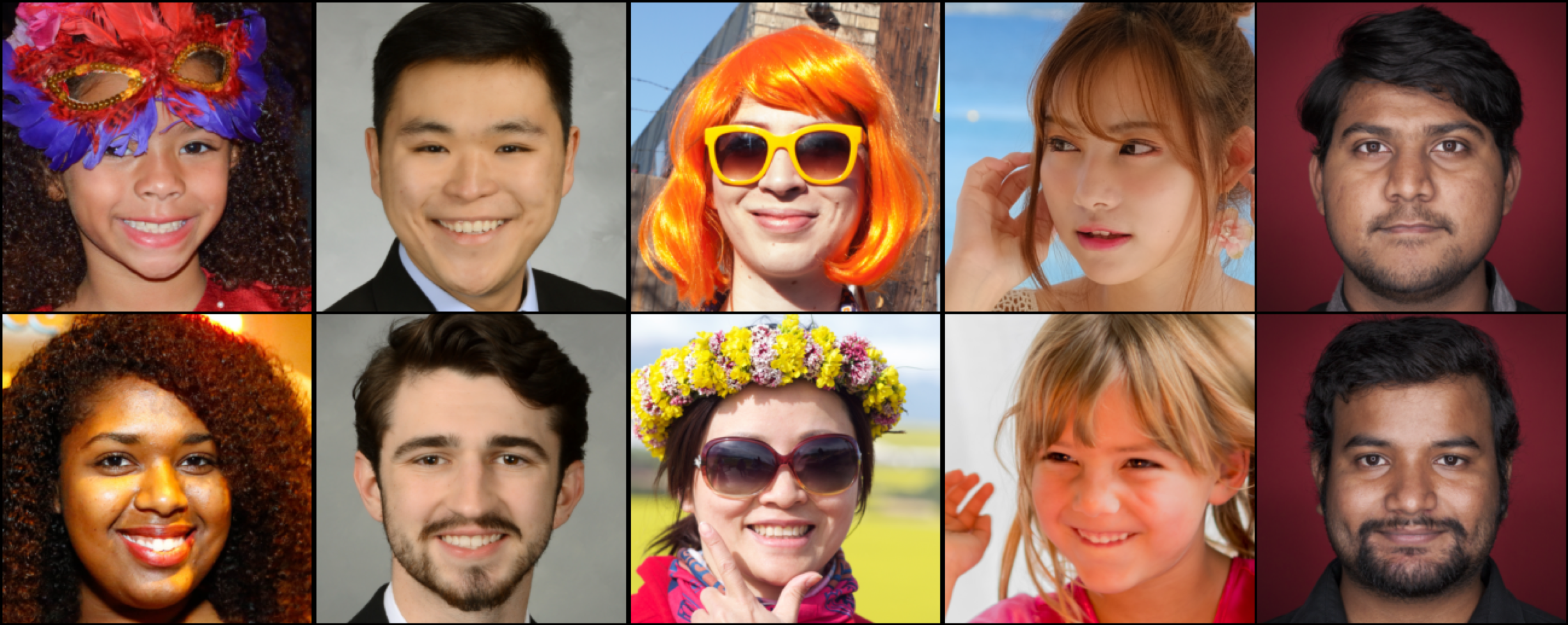}
  \caption{Example closest pairs from greedy agglomerative clustering on embeddings of FFHQ images.}
  \label{cluster pairs}
\end{figure}

\begin{figure*}[ht]
  \centering
  \includegraphics[width=\linewidth]{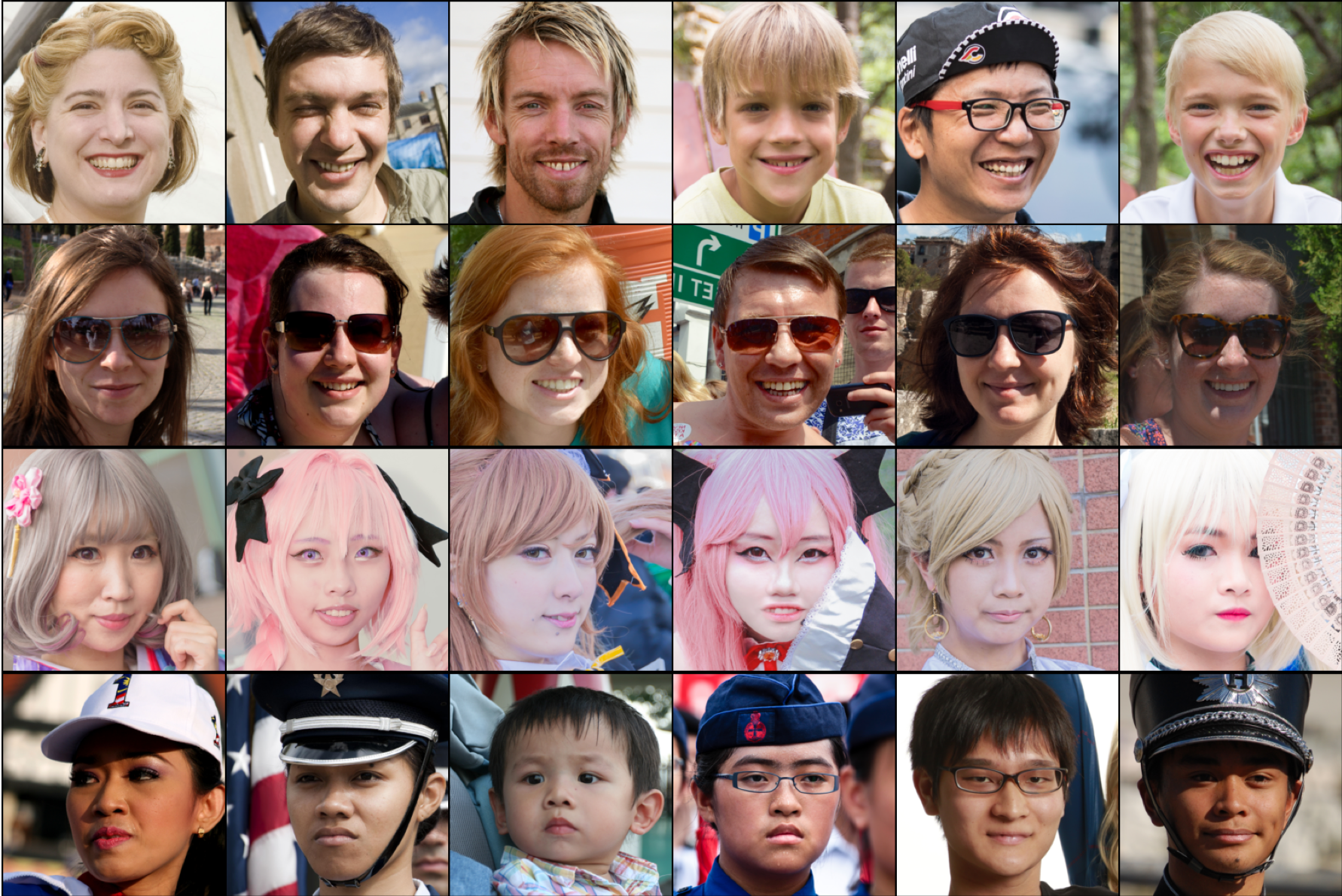}
  \caption{Example clusters from greedy agglomerative clustering on embeddings of FFHQ images.}
  \label{cluster sets}
\end{figure*}

One benefit of the low-dimensional latent space embedding mapped to by $\mathbf{P}$ is that due to its disentangled nature, similar images should have somewhat similar values in certain dimensions of their embedding space. We hypothesize that despite being trained on only images in the distribution of StyleGAN's outputs, $\mathbf{P}$'s learned embedding is able to generalize sets of real images. So we feed 8000 randomly sampled images from the FFHQ dataset through the encoder, and perform greedy agglomerative clustering with Ward's linkage \cite{ward1963hierarchical} on the embedding of the images. This results in sensible clusters that encode various features of faces: gender, glasses, hair style, ethnicity, and overall style. Figure \ref{cluster pairs} demonstrates example closest pairs performed with this agglomerative clustering, and Figure \ref{cluster sets} shows clusters of multiple images.
\section{Conclusion and Future Work}
In this work we propose a method to train an projection network in an unsupervised manner to project images produced by a GAN onto a randomly sampled latent space. We explore using StyleGAN \cite{stylegan} specifically, and demonstrate our results on super-resolution of images produced by the generator. We attempt to extend the projection network to images not produced by the generator (e.g. from the FFHQ dataset), albeit unsuccessfully. However, we perform agglomerative clustering on the latent encoding of images produced by the projection network, which results in semantically meaningful clusters that encode various traits across different people. We believe this work can be extended to perform ``semantic photoshopping'': eventually being able to change various, individual aspects of input images by perturbing the disentangled latent space in the desired directions.
\subsection*{Acknowledgements}
We would like to thank Professors Alex Smola and Mu Li for providing guidance and funding this project.

\end{document}